\documentclass[10.9pt,a4paper]{article}
\usepackage[margin=1in]{geometry}
\usepackage{authblk}
\usepackage{setspace}
\usepackage{amssymb}
\usepackage{amsmath}
\usepackage{tablefootnote}
\usepackage{wrapfig}
\usepackage{multirow}
\usepackage{tabu}
\usepackage{epsfig}
\usepackage{epstopdf}
\usepackage{graphicx, subfigure}
\usepackage{adjustbox}
\usepackage[toc,page]{appendix}
\usepackage{apacite}
\let\cite\shortcite

\newcommand{\qed}{\nobreak \ifvmode \relax \else
	\ifdim\lastskip<1.5em \hskip-\lastskip
	\hskip1.5em plus0em minus0.5em \fi \nobreak
	\vrule height0.75em width0.5em depth0.25em\fi}

\usepackage{xcolor,soul}

\usepackage[bordercolor=white,backgroundcolor=gray!30,linecolor=black,colorinlistoftodos]{todonotes}

\newcommand{\revise}[1]{\textcolor{blue}{#1}}

\usepackage{tikz}
\usetikzlibrary{arrows,calc}
\tikzset{
	>=stealth',
	help lines/.style={dashed, thick},
	axis/.style={<->},
	important line/.style={thick},
	connection/.style={thick, dotted},
}

\graphicspath{{./Figures/}}
\usepackage{amsmath}
\usepackage[algo2e]{algorithm2e}
\usepackage{subfigure}
\usepackage{amsmath}
\usepackage{algorithm}
\usepackage[noend]{algpseudocode}
\usepackage{soul}

\title{Self-Expressive Subspace Clustering to Recognize Motion Dynamics for Chronic Ankle Instability}
\author[1]{Shaodi Qian}
\author[2]{Sheng-Che Yen}
\author[2]{Eric Folmar}
\author[1]{Chun-An Chou\thanks{Corresponding author.}}

\affil[1]{Department of Mechanical \& Industrial Engineering, Northeastern University, Boston, MA 02115}
\affil[2]{Departments of Physical Therapy, Movement \& Rehabilitation Science, Northeastern University, Boston, MA 02115}

\providecommand{\keywords}[1]{\textbf{\textit{Keywords:}} #1}

\begin{document}

\maketitle

\begin{abstract}
Ankle sprains and instability are major public health concerns. Up to 70\% of individuals do not fully recover from single ankle sprains and eventually develop chronic ankle instability (CAI). The diagnosis of CAI has been mainly based on self-report rather than objective biomechanical measures. The goal of this study is to quantitatively recognize the motion patterns of a multi-joint coordinate system using gait data of bilateral hip, knee, and ankle joints, and further distinguish CAI from control cohorts. We propose an analytic framework, where the concept of subspace clustering is applied to characterize the dynamic gait patterns in a lower dimensional subspace from an inter-dependent network of multiply joints. A support vector machine model is built to validate the learned measures compared to traditional statistical measures in a leave-one-subject-out cross validation. The experimental results showed $>$70\% classification accuracy on average for the dataset of 47 subjects (24 with CAI and 23 controls) recruited to examine in our designed experiment. It is found that CAI can be observed from other joints (e.g., hips) significantly, which reflects the fact that there exists inter-dependency in the multi-joint coordinate system. The proposed framework presents a potential to support clinical decisions using quantitative measures during diagnosis, treatment, rehabilitation of gait abnormality caused by physical injuries (e.g., ankle sprains in this study) or even central nervous system disorders.
\end{abstract}

\section{Introduction}
\label{sec1}
 The most common type of ankle injuries is lateral ankle sprains where individuals roll the foot inward and damage the lateral ankle structures \cite{garrick1977frequency}. 
 It is reported that up to 70\% of individuals may eventually develop chronic ankle instability (CAI) following a significant ankle sprain \cite{anandacoomarasamy2005long}. CAI is characterized by persistent pain and swelling, episodes of ankle giving way, and recurrent ankle sprains \cite{hertel2002functional}. Individuals with CAI have been reported to have diminished physical activity \cite{hubbard2015physical} and lower quality of life \cite{houston2015patient}. More problematically, emerging evidence has linked CAI to future development of irreversible, posttraumatic ankle osteoarthritis \cite{golditz2014functional}. The high prevalence and associated medical costs make CAI a significant health concern \cite{soboroff1984benefits}.  

Despite the clinical significance of CAI, the etiology of this medical problem remains unclear. It has been hypothesized that CAI may be caused by peripheral issues such as reduced ankle proprioception \cite{garn1988kinesthetic, forkin1996evaluation} and weak peroneal muscles \cite{bosien1955residual, tropp1986pronator}. However, some counter-evidence has shown that individuals with CAI do not necessarily have reduced proprioception \cite{refshauge2000effect} and ankle strength deficits are not highly correlated with chronic ankle instability \cite{kaminski2002factors}. More recently, researchers started questioning if central control issues also contribute to CAI \cite{hass2010chronic}. A main reason was that individuals with CAI often show deviations in walking and running gaits \cite{monaghan2006ankle, delahunt2006altered}.

Human locomotion requires the control from the central level. Theories such as the central pattern generator have been proposed to explain how the central nervous system controls gait \cite{duysens1998neural, dimitrijevic1998evidence}. Previous studies have found increased ankle inversion \cite{monaghan2006ankle, delahunt2006altered}, increased ankle plantarflexion \cite{chinn2013ankle, drewes2009dorsiflexion}, and altered spatial and temporal parameters \cite{gigi2015deviations} during walking and/or running in individuals with CAI. These findings suggest that the central control of gait is altered in these patients, and support that CAI is not only a peripheral issue but also a central issue. 

Gait represents a complex control problem in which redundant degrees of freedom must be constrained and coordinated to create a smooth pattern \cite{van2005variability}. Thus, examining inter-joint or inter-segment coordination may be able to generate further insight into the central control issue in individuals with CAI. A review of literature revealed that most gait studies on CAI focused on examining individual joints rather than coordination \cite{moisan2017effects}. Few studies examined if CAI is associated with coordination change during gait, but they only looked at coordination between two segments \cite{herb2014shank, drewes2009altered} or two joints \cite{yen2017hip}, e.g., ankle-ankle or hip-ankle. A possible reason may be that the current measurements for coordination, such as vector coding and continuous relative phase, can only quantify the coordination between two body components \cite{hamill2000issues}.

Owing to a significant advancement of sensor technologies, modeling and analysis of high-volume, high-resolution time series data demonstrate successes in various applied domains of healthcare, security, and system monitoring in data-rich environments. \cite{Phinyomark2018,gowsikhaa2014automated}. In particular, gait data analysis is focused on pattern recognition of gait abnormality caused by physical injuries (e.g., ankle sprains \cite{punt2015gait}) or central nervous system disorders (e.g., Parkinson's disease (PD) \cite{pirker2017gait, tucker2015data, duhamel2006functional} and stroke \cite{marrocco2016knee}). 
The essential idea of these studies attempted to extract statistical measures from motion time series, and examine the significance of extracted features using statistical tests (e.g., $t$-test or ANOVA) or statistical learning methods. However, a challenging task is to find discriminant inter-dependent network patterns, instead of independent features, from a multi-variate, high-dimensional gait data space. Most recent researches investigated new quantitative methods for gait pattern recognition of central nervous system disorders, but not of disabilities caused by physical injuries. In this paper, we present the research to specifically model a chronic motion disability problem (i.e., CAI) and identify discriminant patterns based on the self-expression of multi-variate gait data.

In this study, the overarching goal is to characterize critical information (measurements) that can summarize and/or represent the motion behaviors of individuals with CAI using gait kinematic data for a multi-joint coordinate system of bilateral hip, knee, and ankle joints. During our designed running experiments, biosensors were placed to record the locomotions of sagittal, frontal, and transverse planes. We present the multi-joint coordinate system as an inter-connected network as it is hypothesized that there exists inter-dependency between hip, knee, and ankle joints. We then propose an analytic framework that first extracts the network patterns of subspaces learned from multi-variate kinematic data during running, and then develop a decision model using support vector machine (SVM) to validate and distinguish individuals with CAI from control subjects. 

The organization of this paper is as follows. In Section \ref{sec2}, we review relevant research for clinical ankle sprains diagnosis and assessment, followed by a review of quantitative methods for functional data analysis. In Section \ref{sec3}, we describe data acquisition of our designed experiment and then present the proposed analytic framework that involves both unsupervised learning and supervised learning of mutli-variate gait data. In Section \ref{sec4}, computational results are demonstrated and compared between conventional data extraction and our method. Finally, the conclusion and future work are addressed in Section \ref{sec5}.

\section{Related Work}
\label{sec2}

\subsection{Clinical Diagnosis and Assessment of Ankle Sprains}
Traditional clinical assessment for CAI is mainly based on self report. For example, Hiller et al. developed Cumberland Ankle Instability Tool (CAIT) to assess subjective symptoms of CAI \cite{hiller2006cumberland}. Martin et al. developed a self-report survey called The Foot and Ankle Ability Measure (FAAM) that has been widely used in CAI assessment \cite{martin2005evidence}. Another self-report assessment for CAI is the Ankle Instability Instrument developed by Docherty et al. \cite{docherty2006development}. These tools have been recommended by he International Ankle Consortium to determine if an individual has CAI \cite{gribble2014selection}. However, self-report answers may be subject to a number of biases (e.g., social desirability bias) that could affect the reliability and accuracy of the measures. To address this problem, these self-reported tools should be used in conjunction with objective measures.

Previous studies have used lab tools to objectively measure ankle proprioception, balance control, ankle muscle strength, and the reaction time to ankle inversion perturbation in individuals with CAI, but the results were often inconsistent \cite{holmes2009treatment, gutierrez2009neuromuscular}. For example, some studies reported that individuals with CAI have delayed reaction time to inversion perturbation \cite{konradsen1990ankle, karlsson1992effect}, but others did not find similar results \cite{ebig1997effect}. Recently, more studies have examined walking and/or running patterns in individuals with CAI, and the results have suggested that CAI may be associated with deviations in these patterns \cite{monaghan2006ankle, delahunt2006altered, chinn2013ankle, drewes2009dorsiflexion, gigi2015deviations, yen2017hip}. Walking and/or running could be used as a motor task to objectively differentiate control subjects and those with CAI.

Existing gait research on individuals with CAI often focus on examining the ankle kinematics \cite{monaghan2006ankle, delahunt2006altered, chinn2013ankle}. However, gait is a complex motor task that involves all leg joints. Inter-joint coordination could be a parameter used to determine if an individual develops CAI. For example, Yen et al. used a vector coding method to examine hip-ankle coupling during walking, and found that the coupling pattern was different between control subjects and those with CAI \cite{yen2017hip}. A weakness of this study was that the coordination measure was limited to hip-ankle coupling in the affected side in the frontal plane, and deviations may exist in other couplings that were not measured. This weakness was due to limitation in current methods to quantify a coordination pattern. Predominate methods such as vector coding and continuous relative phase can only quantify the coupling between two kinematic trajectories \cite{hamill2000issues}.

\subsection{Pattern Recognition and Analysis of Human Motion Data}

Pattern recognition and analysis of kinematic/gait data \revise{aim} to discover the insights into human motion behaviors through statistical and machine learning methods; recent research studies were surveyed in \cite{Phinyomark2018}. While statistical methods including $t$-test, ANOVA and covariance analysis were used to analyze and test the coupling or grouping across subjects based on statistic measurements for various applications of kinematic and/or physiological data \cite{park2017functional}, advanced techniques of functional data analysis and machine learning demonstrated more promising results in regard to high-dimensional multi-variate gait data \cite{park2017functional}. In various health related studies, supervised learning methods, such as SVM, $k$-nearest neighbors (KNN), linear discriminative analysis (LDA), neural network (NN), were employed for predicting or classifying in between of target and control cohorts, and usually combined with dimensionality reduction approaches, such as principle component analysis (PCA) to discover information from a high-dimensional space \cite{deluzio2007biomechanical,coffey2011common,Fukuchi2011,Eskofier2012,Andrade2013,Phinyomark2014,janidarmian2015analysis,derlatka2015ensemble,tucker2015data,Watari2016,Phinyomark2016,rida2016human}. In addition, some researches were focused on modeling motion dynamics of multi-variate kinematic data using hidden Markov model \cite{MANNINI2012657} and Bayesian network \cite{Moon2008} for gait pattern recognition.

For dimensionality reduction of multi-variate time series data, sparse subspace learning (SSL) or sparse subspace clustering (SSC) is particularly advantageous for reducing the effect of noise while segmenting data in the original space into (multiple) low-dimensional subspaces associated with different groups or functions \cite{noisy,bahadori2015functional}. The idea is to learn an affinity matrix from the original data using sparse or low-rank minimization techniques, and subsequently apply spectral clustering to this affinity matrix to find independent subspaces \cite{sparse}. This subspace clustering method is centered around a self-expressive assumption that a sample point (or times series in a channel) can be linearly represented by other sample points (or channels) in a certain subspace. Furthermore, more efficient algorithms were proposed to solve for the affinity matrix \cite{structured}. In our study, we adopt the same self-expressive assumption as the previous works for dynamic data modeling, and solve for the affinity matrix using an efficient algorithm, called alternating direction method of multipliers algorithm (ADMM) \cite{boyd2011distributed}. It can compute matrix factorization for multiple time series simultaneously. In another work, the self-expression model was solved and further analyzed by other algorithms, e.g., SEED method \cite{dyer2015self} and fast iterative shrinkage-thresholding algorithm \cite{zhang2018dynamic}. The latter research developed a dynamic functional subspace learning method to generate a time-varying correlation matrix for estimating motion dynamics, applied to change point detection \cite{zhang2018dynamic}. A major advantage is that large fluctuations are prevented in estimating over time. However, their algorithm is not computationally efficient and computes individual time series one at a time. In a previous study \cite{sparse}, they considered SSC results of geometric interpretation, i.e., principal angles between subspaces, for object detection and clustering in video footage, and face recognition (photo clustering). Meanwhile, SSC concept is found in many applications e.g., image compression \cite{Hong2006}, motion segmentation \cite{Rao2010}, video segmentation \cite{sparse}, social community clustering \cite{Chen2014}, and gene expression profiles clustering \cite{McWilliams2014}. In this paper, we consider the SSC concept in investigating chronic motion disability of multi-variate kinematic data. In our analytical framework, with the self-expressive assumption, the SSC result (sparse subspace matrix) is obtained by optimizing the objective function using the ADMM algorithm and used as discriminative features for CAI pattern classification.

\section{Materials and Methods}
\label{sec3}

\subsection{Data Acquisition and Processing}
\label{subsec1}

In this study, we acquired human motion data in running mode from 47 subjects in total, among which 24 subjects with CAI and 23 control subjects. There are 36 females (18 subjects with CAI and 18 controls) and 11 males (6 subjects with CAI and 5 controls). Subjects were recruited from the Northeastern University community. Screening exams were conducted for all potential subjects to determine their suitability for the study based on the inclusion and exclusion criteria. We followed the selection standards endorsed by the International Ankle Consortium to set the inclusion criteria for CAI \cite{gribble2014selection}. All subjects with CAI had a signiﬁcant ankle sprain at least one year before being enrolled in this study. In addition, they scored 24 or lower in CAIT and had more than one episode of ankle giving ways in the past 6 months.

\begin{figure}[htbp]
    \centering
    \includegraphics[scale=0.45]{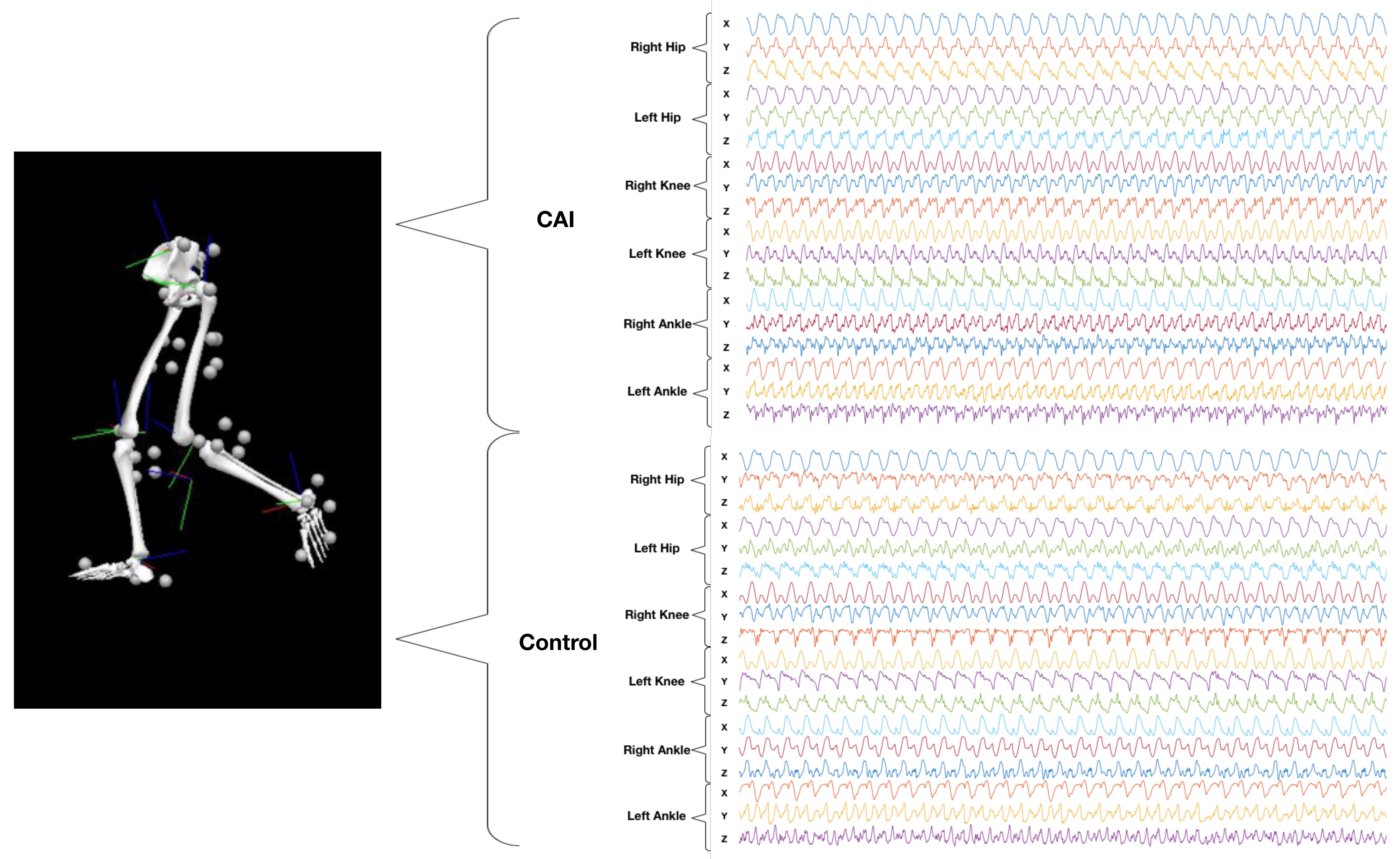}
    \vspace{-0.2in}
    \caption{A representation of kinematic data collection from a 7-camera 3D motion capture system. The upper panel presents eighteen time series from bilateral hip, knee, and ankle joints of a subject with CAI and the lower panel presents eighteen time series from the same joints of a control subject.}
    \label{fig:motion}
\end{figure}

All qualified subjects were asked to participate in a single session, in which they were asked to run on a treadmill for one minute. Their running performances and behaviors were captured by a 7-camera 3D motion capture system (Qualisys AB, Sweden). Reflective markers were placed on the major bony landmarks of the pelvis and left and right legs in order to create biomechanical models of bilateral hip, knee, and ankle. The recorded marker data were then analyzed using Visual3D (C-Motion, MD). The resultant variables are bilateral hip, knee, and ankle position changes in each of the sagittal, frontal, and transverse planes over time. Limited by the number of subjects, we consider running cycles (unit cycle) in each subject, instead of individual subjects, as samples. Each joint trajectory was segmented into cycles (defined as the interval between two consecutive initial contacts of the same foot) for further analysis. On average, each session contains around forty cycles varied by subject's running posture. Figure \ref{fig:motion} shows the kinematic data collected from the motion capture system for both CAI and control subjects. For each joint, there are three motion time series in x-axis, y-axis, and z-axis presenting sagittal, frontal, and transverse planes. There are eighteen channels in each motion dataset. We define channels 1-3 representing the sagittal, frontal, and transverse planes on right hip, channels 4-6 representing the sagittal, frontal, and transverse planes for left hip, and so on. Since the total number of time points in running cycles varies in different subjects, we applied a linear interpolation method to ensure all cycles' length to 84 time points, which is the longest running cycle in the dataset. Then, motion time series data were detrended and z-scored to avoid variability between subject postures and potential human errors during data collection. As a result, for a subject, there are $\sim$40 matrices, denoted by $y_{i} (t)$ where $i = 1, ..., 18$ and $t = 1, ..., 84$.  

\subsection{Proposed Method}

\begin{figure}[htbp]
    \centering
    \includegraphics[scale=0.45]{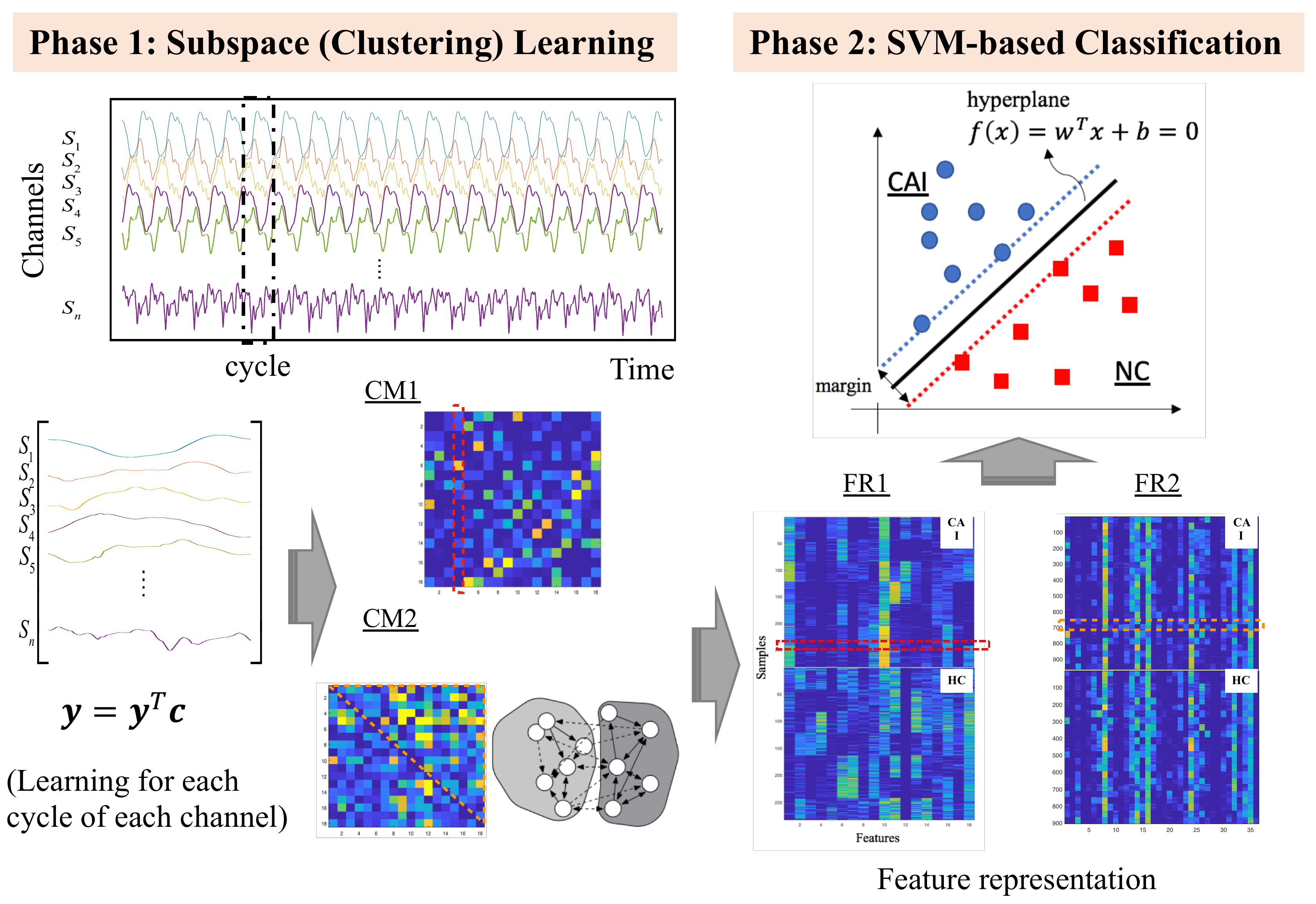}
    \caption{An illustration of our proposed analytic framework. In Phase 1, dynamic motion patterns (i.e., clustering coefficients \textbf{C}) are recognized by sparse subspace clustering for each running cycle to represent the motion behaviors of the multi-joint coordinate system. In Phase 2, a linear SVM-based classification model is trained with clustering coefficients for binary classification tasks between CAI and control cohorts.}
    \label{fig:framework}
\end{figure}

In this section, we describe our proposed analytic framework composed of two phases to recognize the dynamic motion patterns across channels in the multi-joint coordinate system for discriminating and predicting subjects with CAI from a control cohort. Note that in this study, we use a term `channel' to express sagittal, frontal, or transverse plane of a joint. Figure \ref{fig:framework} illustrates the concept of our proposed method. In the first phase, we learn and extract significant clustering (or inter-connection) patterns of segmented motion signals. For each running cycle of individual subjects, motion signals undergo the detrending and z-scoring pre-processing steps. Then, a sparse subspace clustering \cite{sparse} is adopted to obtain clustering coefficients (CCs) to represent the dynamical inter-relationships among channels. In the second phase, we consider a linear SVM-based classification model using the estimated CCs as input features for performing a binary classification task between CAI and control cohorts. Both channel-based and network-based modeling strategies (CM1 and CM2) are proposed. First, for a channel, we use its clustering coefficients (i.e., a column vector in the CC matrix) as input features, highlighted in red rectangular in Figure \ref{fig:framework}, which present the correlation of other channels to it due to the self-expression assumption. Second, we compute a symmetric CC matrix. Then a clustering analysis is further applied to find a more representative network, and use all CCs (in the triangular CC matrix), highlighted in yellow rectangular, as input features. The detailed methodology is presented as follows.


\subsubsection{Self-Expressive Sparse Subspace Learning}
\label{sec41}
Let us consider a set of multi-variate motion time series $\mathbf{Y}(t) = \{y_{1}(t),y_{2}(t), ..., y_{n}(t)\}$, where $n$ = 18, in a multi-joint coordinate system. It is considered as a dynamic system of multiple units where each unit linearly or nonlinearly interact with each other at different connectivity degree and can be presented in certain subspaces $g_m \in \mathbf{G}$, where $m$ is the number of subspaces. In this study, it is assumed that each motion time series is self-expressive and presented by a linear combination of some other motion time series and they share the same subspace. In other words, each motion time series can be represented by the basis functions in a subspace shared with other motion time series, which are highly correlated with each other while those in different subspaces have no or weak correlations. Note that the number of subspaces ($m$) need not to be pre-determined and will be determined by solving a self-expressive sparse subspace clustering problem.



The mathematical formulation of self-expressive model is presented as follows:
\begin{equation}
\textbf{y}_i = \mathbf{Y} \textbf{c}_i, \ \textbf{c}_i^T\textbf{1} = 1, \ c_{ii} = 0, \ \forall \ i = 1, \dots, n,
\end{equation}
where $\textbf{Y}$ is the original time series in $n$ channels and $\textbf{c}_i = [c_{i1}, c_{i2}, \dots, c_{in}]$ is a coefficient vector. 
To find a lower-dimensional subspace shared by original channels, we aim to minimize the coefficient values. The optimization problem for SSC is then formulated as follows:
\begin{equation}
 \min_{\textbf{c}_i}\quad \lVert \textbf{c}_{i} \rVert \quad s.t.\quad \textbf{y}_i = \textbf{Y}\textbf{c}_i, \ \textbf{c}_i^T\textbf{1} = 1, \ \quad c_{ii}=0.
\end{equation}
After solving this minimization problem, a sparse solution of $\textbf{c}_i$ is obtained, where the non-zero elements $c_{ij} \neq 0$ express the coefficients of channel $j$ to channel $i$. Furthermore, we extend to solve the whole system by introducing a coefficient matrix \textbf{C} = $[\textbf{c}_{1}, \textbf{c}_{2}, \dots, \textbf{c}_{n}]$. The extended optimization problem can be written as
\begin{equation}
\label{original}
\min_\textbf{C}\quad \lVert \textbf{C} \rVert \quad s.t.\quad \textbf{Y} = \textbf{YC}, \quad diag(\textbf{C}) = 0, \ \textbf{C}^T \textbf{1} = \textbf{1}.
\end{equation}
Since noises usually appear in real data, we assume $Y_{i}(t)=X_{i}(t)+\epsilon_{i}(t),$ where $X_{i}(t)$ represents the real signals and $\epsilon_{i}$ is independent and identically distributed noise with zero mean. We rewrite the model in Equation (\ref{original}) as follows:
\begin{equation}
\label{E}
\min_{\textbf{C,E}} \quad \lVert \textbf{C} \rVert + \lambda \lVert \textbf{E} \rVert \revise{^2} \quad s.t.\quad \textbf{Y} = \textbf{YC} + \textbf{E},\quad diag(\textbf{C})=0, \ \textbf{C}^T \textbf{1} = \textbf{1}. 
\end{equation}
In the objective function, we still minimize the coefficients and add a penalty term to minimize the fitting errors, where $\lambda$ is the penalty parameter, subject to the same constraints. By solving the model Equation (\ref{E}), we reveal corresponding subspaces and obtain a sparse representation of CC matrix $\textbf{C}$, in which the nonzero elements represent the channels from the same subspace. Ideally, if both coefficients $c_{ij}$ and $c_{ji}$ are nonzero, then we consider channel $i$ and $j$ share the same subspace.


In real-life cases, however, inter-dependency and/or interactions are not usually linear explicitly. We herein introduce a kernel trick, a commonly used technique in pattern recognition, to map the original motion time series in a non-linear dimensional space, where subspace can be linearly expressed.
Then, we can rewrite the optimization problem for kernel sparse subspace clustering (KSSC) as follows \cite{kernel1}:
\begin{equation}
\label{kernel}
    \min_\textbf{C}\quad \lVert \textbf{C} \rVert + \lambda \lVert \psi (\textbf{Y})-\psi (\textbf{Y})\textbf{C} \rVert^2 \quad
    s.t. \quad diag(\textbf{C})=0,\quad \textbf{C}^T \textbf{1} = \textbf{1},
\end{equation}
where $\psi(\cdot)$ is a mapping function and $\psi(\cdot) = \{ \psi(\textbf{y}_i), \psi(\textbf{y}_2), \dots, \psi(\textbf{y}_n) \}$. A positive semi-definite kernel matrix is defined as $\kappa(Y,Y)_{ij} = \langle \psi(\mathbf{Y}_i) ,\psi(\mathbf{Y}_j) \rangle$ and Gaussian kernel is used in our study. It is noted that the inter-dependency/interactions are still linear in the mapped subspace.

\subsubsection{ADMM Solution Approach} \label{admm}
To solve both SSC and KSSC optimization problems, we adopt a well-developed alternating direction method of multipliers (ADMM) algorithm \cite{boyd2011distributed}. We present the solution process for KSSC optimization problem and it can be done for SSC optimization problem in a similar way. We first rewrite the model in Equation (\ref{kernel}) by introducing an auxiliary matrix $\textbf{A} \in R^{N\times N}$
as follows \cite{sparse}:
\begin{equation}
\begin{split}
\label{aff}
& \min_{\textbf{A,C}} \ \lVert \textbf{C} \rVert+ \lambda \lVert \psi (\textbf{Y}) - \psi (\textbf{Y})\textbf{A}\rVert^2\\
& s.t. \quad \textbf{A} = \textbf{C}-diag(\textbf{C}), \quad \textbf{A}^T\textbf{1} = \textbf{1}.
\end{split}
\end{equation} 
Then, we add two penalty terms to make the problem convex in terms of variables \textbf{C} and \textbf{A}, where $\frac{\rho}{2}$ is the trade-off parameter between the two terms. The convex optimization problem is expressed as follows:
\begin{equation}
\begin{split}
\label{aff2}
& \min_{\textbf{A,C}} \ \lVert \textbf{C} \rVert+ \lambda \lVert \psi (\textbf{Y}) - \psi (\textbf{Y})\textbf{A}\rVert^2 + \frac{\rho}{2}\lVert \textbf{A} - \textbf{C} + diag(\textbf{C})\rVert^2+\frac{\rho}{2}\lVert \textbf{A}^T\textbf{1}-\textbf{1}\rVert^2\\
& s.t. \quad \textbf{A} = \textbf{C}-diag(\textbf{C}), \quad \textbf{A}^T\textbf{1} = \textbf{1}.
\end{split}
\end{equation} 
Note that the models in Equation (\ref{aff}) and Equation (\ref{aff2}) share the same solution since all feasible solutions that satisfy all the constraints ensure two penalty terms equal to zero. Furthermore, we derive a new formulation for KSSC optimization problem by applying Lagrange multipliers a vector $\textbf{$\delta$}\in R^N$ and a matrix $\textbf{$\Delta$}\in R^{N \times N}$ as follows:
\begin{equation}
\begin{split}
\label{complete}
\min_{\textbf{A,C}} \  &\lVert \textbf{C}\rVert_1 + \lambda\lVert \psi (\textbf{Y})-\psi (\textbf{Y})\textbf{A}\rVert^2+\frac{\rho}{2}\lVert \textbf{A}-\textbf{C}+diag(\textbf{C})\rVert^2+\frac{\rho}{2}\lVert \textbf{A}^T\textbf{1}-\textbf{1}\rVert^2+\\
&\textbf{$\delta$}^T(\textbf{A}^T*\textbf{1}-\textbf{1})+tr(\textbf{$\Delta$}^T(\textbf{A}-\textbf{C}+diag(\textbf{C}))),
\end{split}
\end{equation}
The ADMM is aimed to blend the decomposability and solving this convex optimization problem by decomposing it into small sub-problems in sequence and solving individual variables one at a time \cite{boyd2011distributed}. The solving procedure iterates until $\textbf{C}$ is converged within an acceptable error or a pre-set maximum iteration is reached. For any iteration $k$:
\begin{enumerate}
    \item Update $\textbf{A}^{k+1}$ by optimizing Equation (\ref{complete}) with respect to $\textbf{A}$ with fixed $\textbf{C}^{k}$, $\delta^{k}$ and $\Delta^{k}$. We obtain $\textbf{A}^{k+1}$ by calculating the following.
$$\left( \lambda_z\textbf{K}_{YY}+\rho \textbf{I}+\rho \textbf{1}\textbf{1}^T \right)\textbf{A}^{k+1}=\lambda_z\textbf{K}_{YY}+\rho(\textbf{1}\textbf{1}^T+\textbf{C}^k)-\textbf{1}\textbf{$\delta$}^{k^T}-\textbf{$\Delta$}^k$$
\item Update $\textbf{C}^{k+1}$ by optimizing Equation (\ref{complete}) with respect to $\textbf{C}$ with updated $\textbf{A}^{k+1}$ and fixed $\delta^{k}$ and $\Delta^{k}$. We obtain $$\textbf{C}^{k+1}=\textbf{D}-diag(\textbf{D}),\quad \textbf{D}=T_{1/\rho}(\textbf{A}^{k+1}+\textbf{$\Delta$}^k/\rho),$$
Note that $T_\eta(\cdot)$ is the shrinkage-thresholding operator applied on each element in matrix and is defined as $T_\eta(\upsilon)=(|\upsilon|)_+sgn(\upsilon)$. Operator $(\cdot)_+$ returns its argument if non-negative, and returns zero otherwise.
\item Update $\textbf{E}^{k+1}$ by optimizing Equation (\ref{complete}) with respect to $\textbf{E}$ with updated $\textbf{C}^{k+1}$ and $\textbf{A}^{k+1}$, and fixed $\delta^{k}$ and $\Delta^{k}$. $$\textbf{E}^{k+1}=T_{\frac{\lambda_e}{\lambda_z}}(\textbf{Y}-\textbf{YA}^{k+1})$$
\item Updating $\delta^{k+1}$ and $\Delta^{k+1}$ with updated $\textbf{C}^{k+1}, \textbf{A}^{k+1}$. $$\textbf{$\Delta$}^{k+1}=\textbf{$\Delta$}^k+\rho(\textbf{A}^{k+1}-\textbf{C}^{k+1})$$
$$\textbf{$\delta$}^{k+1}=\textbf{$\delta$}^k+\rho(\textbf{A}^{{k+1}^T}\textbf{1}-\textbf{1})$$
\end{enumerate}

The pseudo-code is summarized in Algorithm 1.
\begin{algorithm}[ht]
\label{pseudo}
 \caption{ADMM} \label{alg1} 
 \KwData{$K_{YY}$, hyper-parameters: penalties, Max-Iter, Threshold}
 \KwResult{Sparse subspace matrix \textbf{C}}
 \textbf{Initialization:} $\textbf{A}$, $\textbf{C}$, $\textbf{E}$, \textbf{$\Delta$} and \textbf{$\delta$} = \bf{0}\;\\
 \While{$errors \le threshold$ or reach Max-Iter}{
    1. Solve following equation for $\textbf{A}^{k+1}$:\\
    \centerline{$\left( \lambda_z\textbf{K}_{YY}+\rho \textbf{I}+\rho \textbf{1}\textbf{1}^T \right)\textbf{A}^{k+1}=\lambda_z\textbf{K}_{YY}+\rho(\textbf{1}\textbf{1}^T+\textbf{C}^i)-\textbf{1}\textbf{$\delta$}^{i^T}-\textbf{$\Delta$}^i$}\\
    2. Update \textbf{C} by:\\ \centerline{$\textbf{C}^{k+1}=\textbf{D}-diag(\textbf{D}),\quad where \quad \textbf{D}=T_{1/\rho}(\textbf{A}^{k+1}+\textbf{$\Delta$}^i/\rho)$}\\
    3. Update \textbf{E} by: $\textbf{E}^{k+1}=T_{\frac{\lambda_e}{\lambda_z}}(\textbf{Y}-\textbf{YA}^{k+1})$\\
    4. Update \textbf{$\Delta$} by:\\ \centerline{$\textbf{$\Delta$}^{k+1}=\textbf{$\Delta$}^i+\rho(\textbf{A}^{k+1}-\textbf{C}^{k+1})$}\\
    \centerline{$\textbf{$\delta$}^{k+1}=\textbf{$\delta$}^i+\rho(\textbf{A}^{{k+1}^T}\textbf{1}-\textbf{1})$}
    5. $k=k+1$\\
    }
\end{algorithm}

For optimizing hyper-parameters $\lambda$ and $\rho$, we used grid search method to find the best settings in terms of the highest cross-validation accuracy: $\lambda = 0.015$ and $\rho$ = 800 for SSC problem, and $\lambda = 60$ and $\rho = 2500$ for KSSC problem.


\subsubsection{Feature Extraction and Representation}
\label{DBSCAN}

For input features used in a pattern classification task in the next section, we propose two strategies such that motion pattern is recognized and distinguished in the learnt subspace. First, to examine the discriminative power, we consider individual channel-based vector $\mathbf{c}_i$ in the clustering coefficient matrix $\mathbf{C}$ (see \ul{CM1} in Figure \ref{fig:framework}). Each channel-based vector $i$ contains a set of non-zero coefficients of other channels $i' \in I \setminus i$ correlating to channel $i$ in the same subspace. Channel-based vectors from all cycles of all subjects are aggregated to form an input feature set. There are $|I|$ sets of input features to be performed in the classification task separately. 

Second, we calculate an affinity symmetric matrix $\textbf{C}_{s}=|\textbf{C}|+|\textbf{C}|^T$ to present the network (or connectivity) pattern across all channels (see \ul{CM2} in Figure \ref{fig:framework}). We employ a state-of-the-art clustering technique, called density-based spatial clustering of applications with noise (DBSCAN) \cite{dbscan1}, to find clustering subspaces $\mathbf{G} = \{g_1, g_2, ..., g_m\}$, in which each cluster is formed by channels that interact with each other closely and dynamically; that is, connectivities in between are strong. As DBSCAN is aimed to discover clusters of input channels, two parameters are required to set up: $\theta$ is a distance threshold to determine the closeness for the neighborhood of a given sample and $\phi$ is a threshold to limit the number of neighbors to form a cluster. In our study, we consider the Euclidean distance in the feature space. To partition all channels into two parts for all subjects, we set the threshold $\theta$ equal to the median distance since we like to include as many channels as possible in the neighborhood and $\phi$ = 4 for which we generate consistent clustering result across all subjects. DBSCAN starts with an arbitrary channel and includes all other channels that are directly density-reachable as its neighborhood is satisfied with the two parameter thresholds. This process iterates until no new channels being added to this neighborhood. Note that this is computationally fast since there are only 18 channels for clustering. As a result, two major clusters ($m = 2$) are formed. We then perform sparse subspace learning, described in Section \ref{sec41}, on those channels included in individual clusters $g_1$ and $g_2$, respectively. For each cluster, we obtain an updated clustering matrix $\mathbf{C}_{s}$ (see \ul{FR2} in Figure \ref{fig:framework}), where the elements in the upper-triangle part in $\mathbf{C}_{s}$ are referred to as network coefficients in the new subspace.

\subsubsection{Pattern Classification} \label{sec42}
 With the learned sparse clustering matrices $\mathbf{C}$ and $\mathbf{C}_s$ in Section \ref{admm}, we propose to train a SVM-based classification model for performing binary classification tasks between CAI and control cohorts. The objective is to validate the effectiveness of the learned motion patterns for distinguishing CAI subjects from control subjects. SVM is considered as one of robust supervised learning methods in various applied problems of high-dimensional data \cite{hearst1998support}. The idea of SVM is to find a hyperplane with a maximum margin that allows to separate the samples in two classes (i.e., CAI versus control). Let us have training data $\mathbf{x} = \{x_{lk}\}$ as samples $l \in L$ presented by feature set $k \in K$ and binary class $y_l = \{-1, 1\}$ associated with the samples. A hyperplane is defined as 
\begin{equation}
    h(\mathbf{x}) = \mathbf{w}^{T} \mathbf{x} + b,
\end{equation}
where $\mathbf{w}^{T}$ is a weight vector and $b$ is a scalar. A sample is decided to belong to class $y = +1$ if $h(\mathbf{x}) > 0$; otherwise, it belongs to class $y = -1$ if $h(\mathbf{x}) < 0$. To find an optimal hyperplane $h^{*} (\mathbf{x}) = arg \max_{\mathbf{w}, b} \{\frac {1} {\|\mathbf{w}\|}\}$, one can solve a constrained optimization problem to optimize the in-between margin, shown as follows:

\begin{equation}
    \min_\textbf{w} \quad \frac {\|\mathbf{w}\|^2} {2} + \gamma \sum_{l \in L} \xi_l, \quad \mathrm{s.t.} \ y_l (\mathbf{w}^{T} \mathbf{x}_l + b) \geq 1 - \xi_l \ \forall \ l \in L; \quad \mathbf{w}, b \in \Re, \ \xi_l \geq 0, \ \forall \ l \in L.  
\end{equation}
We solve this classification problem using a widely used library LIBSVM \cite{chang2011libsvm} with the default settings. For the sake of interpretation, a linear kernel is used throughout the computational experiments in this study. The optimized weights tell the features importance; the larger positive or negative value is, the more discriminative the feature is. See the resultant example in Figure \ref{svmweight}. The detail will be discussed in the latter section.

\section{Computational Results} 
\label{sec4}

\begin{figure*}[t!]
    \centering
    \includegraphics[width=1\linewidth]{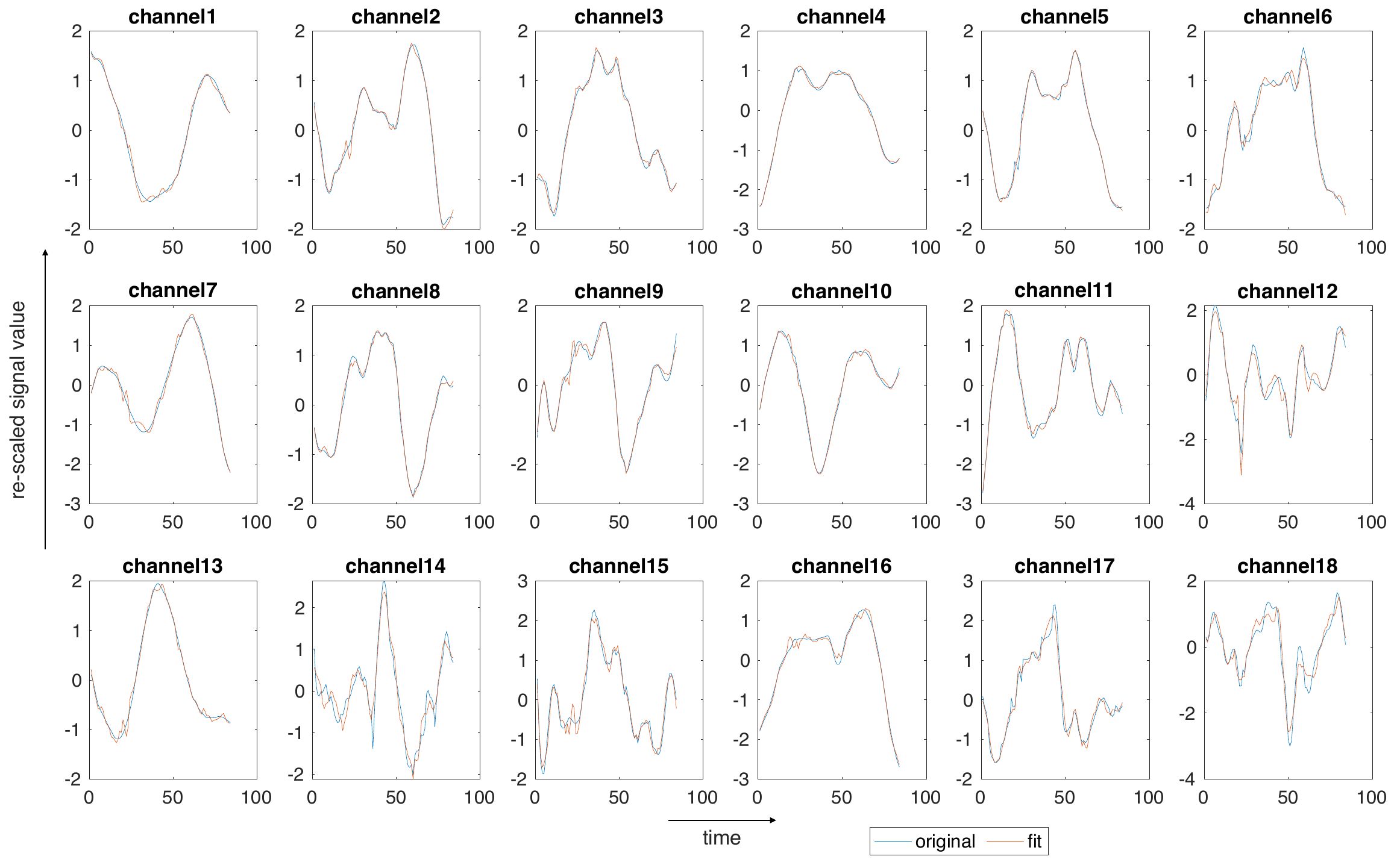}
    \caption{An illustration of the fitted result using the SSC method for one cycle of 18 channels on Subject 01.}
    \label{fit}
\end{figure*}

In this section, for CAI dynamic motion pattern recognition, we perform our proposed subspace clustering (SSC and KSSC) methods compared to traditional methods including statistic features and Pearson correlation, and PCA. We train a linear SVM model with these extracted features for classification in a leave-one-subject-out cross validation schame; each subject of $\sim$40 sample cycles is left for testing iteratively. We also examine the goodness-of-fit of our methods, and obtain R-squared equal to 0.98 on avereage for all subjects. In Figure \ref{fit}, we illustrate the fitted curves (in red) give a good approximation to the original running curves (in blue) for one cycle of all channels in Subject 1. 

\subsection{Results for Channel-based Subspace} 
\label{result}
\begin{figure*}[t!]
    \centering
    \subfigure[SSC]{
        \includegraphics[width=0.7\linewidth]{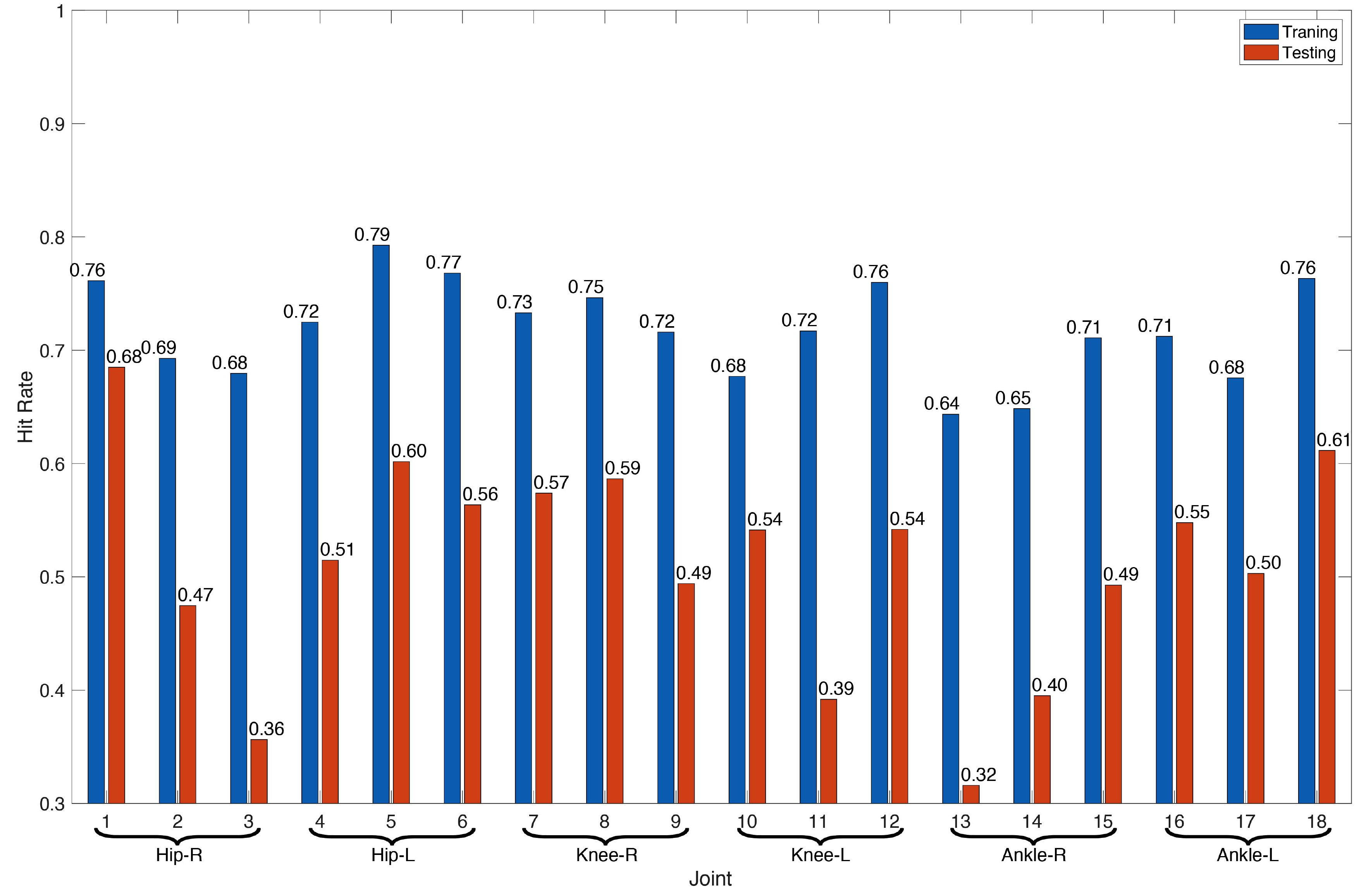}}
    \subfigure[KSSC][KSSC]{
        \includegraphics[width=0.7\linewidth]{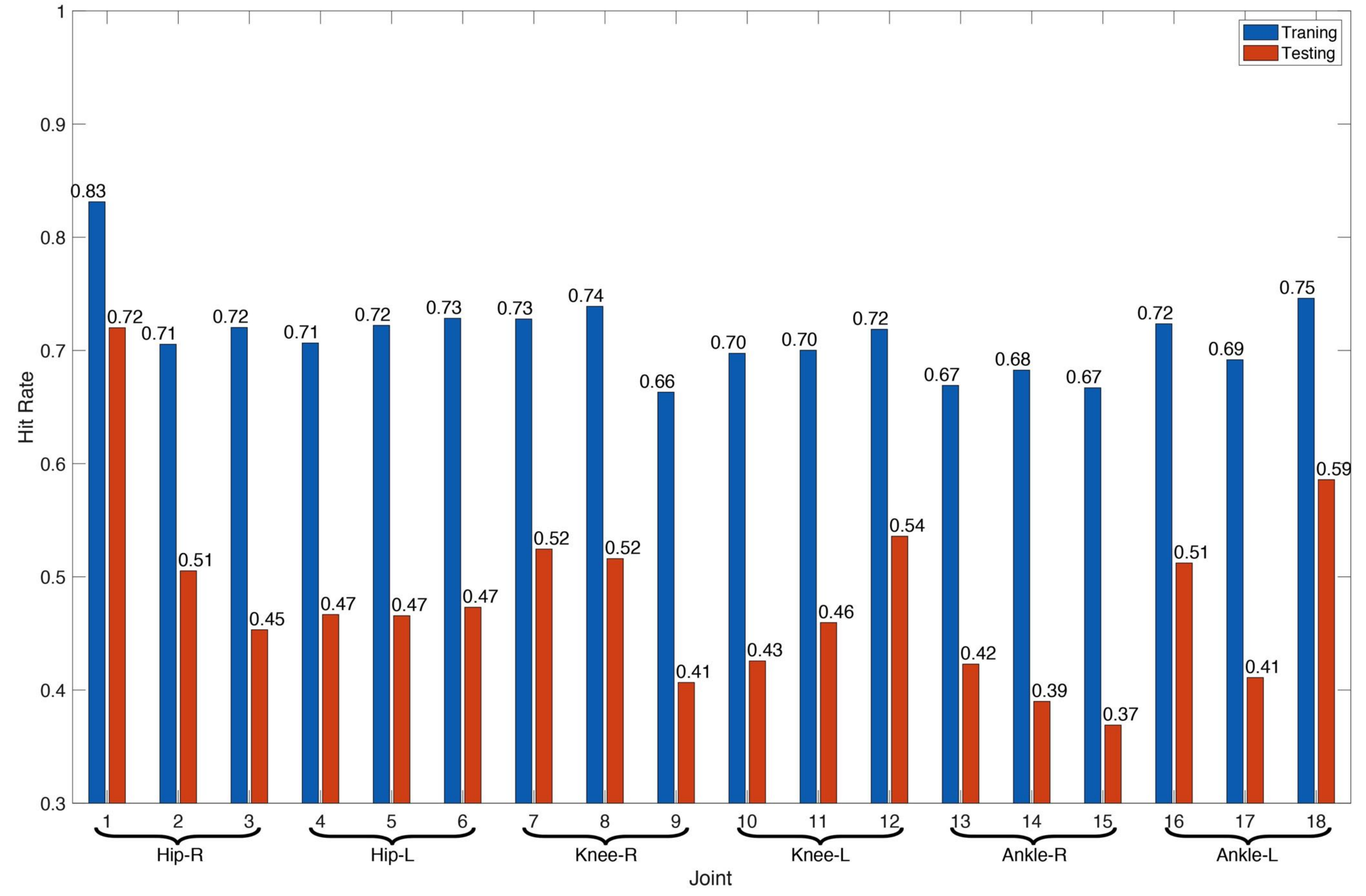}}
    \caption{Hit rates for individual channel-based subspace using SSC and KCCS features, respectively.}
    \label{barplot_hr}
\end{figure*}

\begin{figure*}[t!]
    \centering
    \subfigure[SSC]{
        \includegraphics[width=0.7\linewidth]{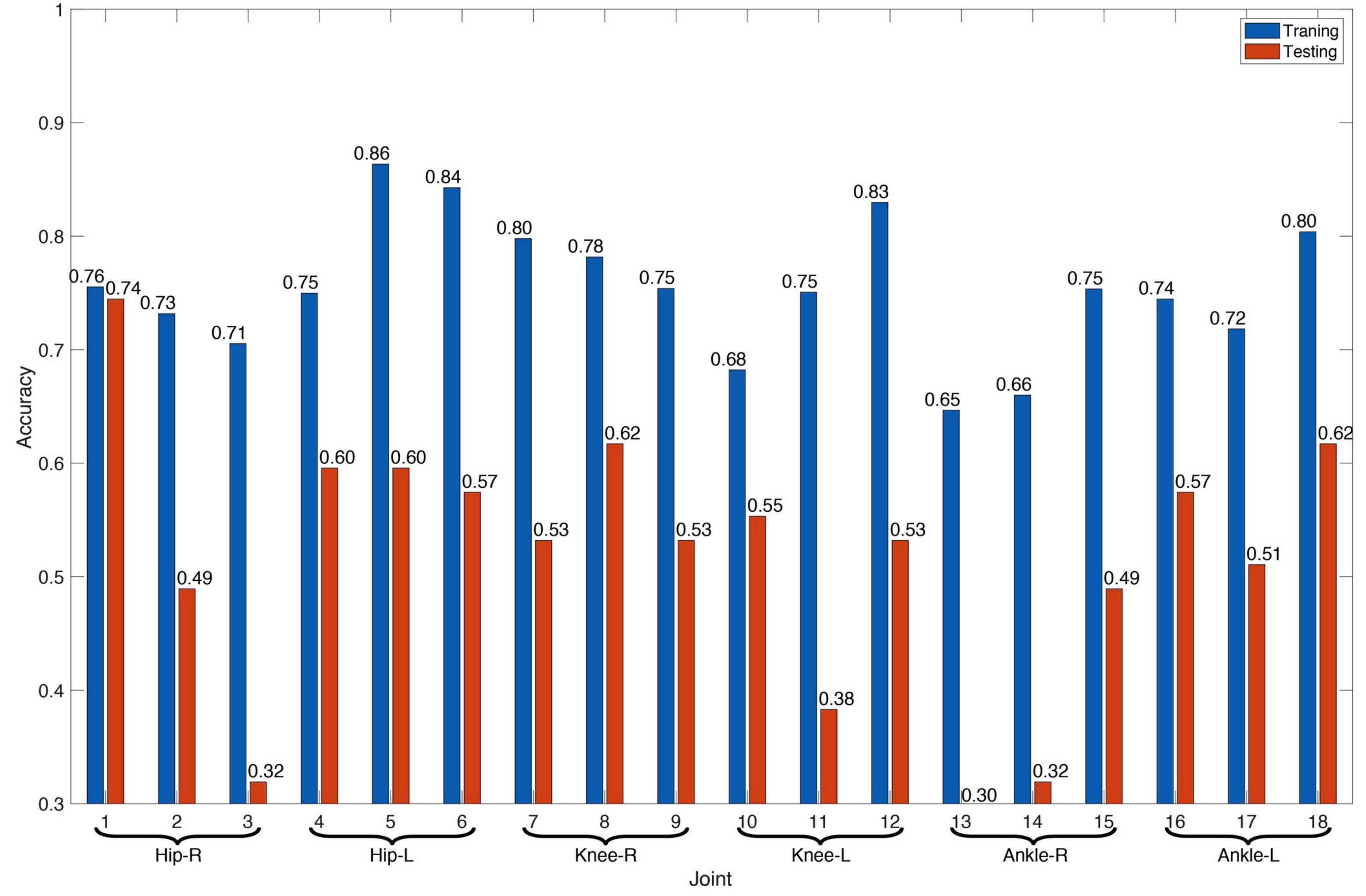}}
    \subfigure[KSSC][KSSC]{
        \includegraphics[width=0.7\linewidth]{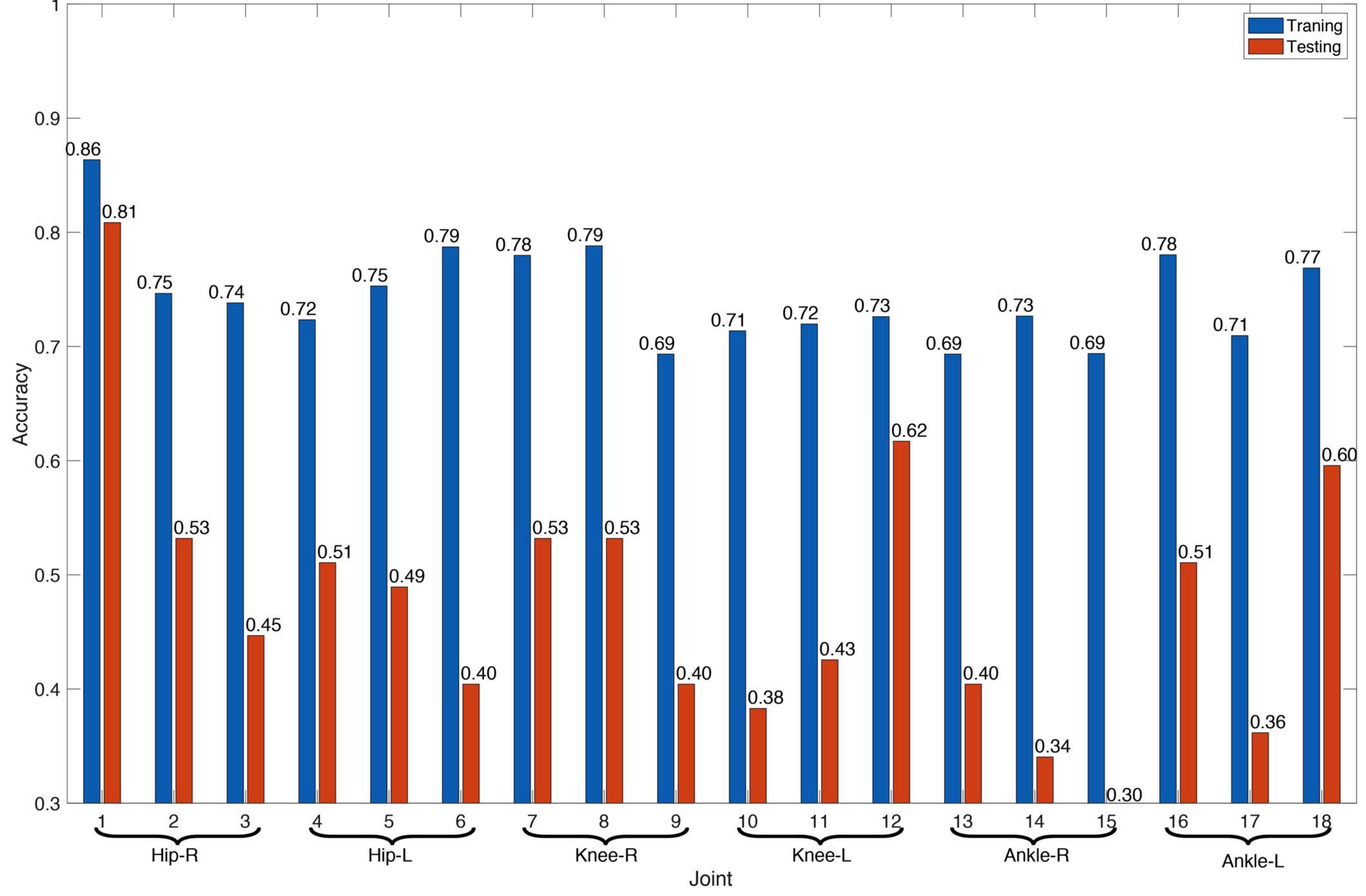}}
    \caption{Majority voting accuracies for individual channel-based vectors using SSC and KCCS features, respectively.}
    \label{barplot}
\end{figure*}

To examine the discriminative power of learnt motion patterns in different subspaces, we perform a classification task for individual learnt subspaces using channel-based vectors in the matrix $\mathbf{C}$. We evaluate using two measures: (1) hit rate HR = $\frac{TP}{TT}$ for individual subjects, where true positive ($TP$) is the number of correctly classified sample cycles (matching true class) and $TT$ is the total number of sample cycles, and (2) majority voting accuracy ($MV\_Accur$). Subject is correctly classified as the percentage of correctly classified sample cycles exceeds 50\%. The results based on the two evaluation metrics are presented in Figures \ref{barplot_hr} and \ref{barplot}, respectively. In Figure \ref{barplot_hr}, for both SSC and KSSC features, we see that the only subspace based on Channel 1 (sagittal plane of right hip) results in $\sim$70\% testing hit rate, followed by the other subspace based on Channel 18 (transverse plane of left ankle). The rest of subspaces based on other channels results in $< 50\%$ testing hit rate. In Figure \ref{barplot}, with SSC features, there is the only one subspace based on Channel 1 (sagittal plane of right hip) resulting in $\sim80\%$ testing $MV\_Accur$. The subspaces based on Channels 3, 11, 13, 14, and 15 result in $< 50\%$ testing $MV\_Accur$, and there exists a relatively larger gap between training and testing accuracies. With KSSC features, there is the only one subspace based on Channel 1 (sagittal plane of right hip) resulting in $> 80\%$ testing $MV\_Accur$ and the rest of subspaces based on other channels results in $< 60\%$ testing $MV\_Accur$.

It is interesting that most CAI subjects can be distinguished from the hip position (Channel 1) rather than ankle position (Channels 13-18). The motion abnormality on knees along has little influence or impact on the running patterns of subjects. It is noted that there a big gap (around 20-40\%) between testing and training accuracies except Channel 1, but is commonly seen in cross-subject validation due to high variation between samples in different subjects. Limited by the number of subjects, it is difficult to eliminate the bias between subjects with different body conditions, such as BMI, gender, dominant and affected leg. It may be nothing related to over-fitting in the classification model building.


\begin{figure*}[t!]
    \centering
    \subfigure[SSC]{
    \includegraphics[width=0.48\textwidth]{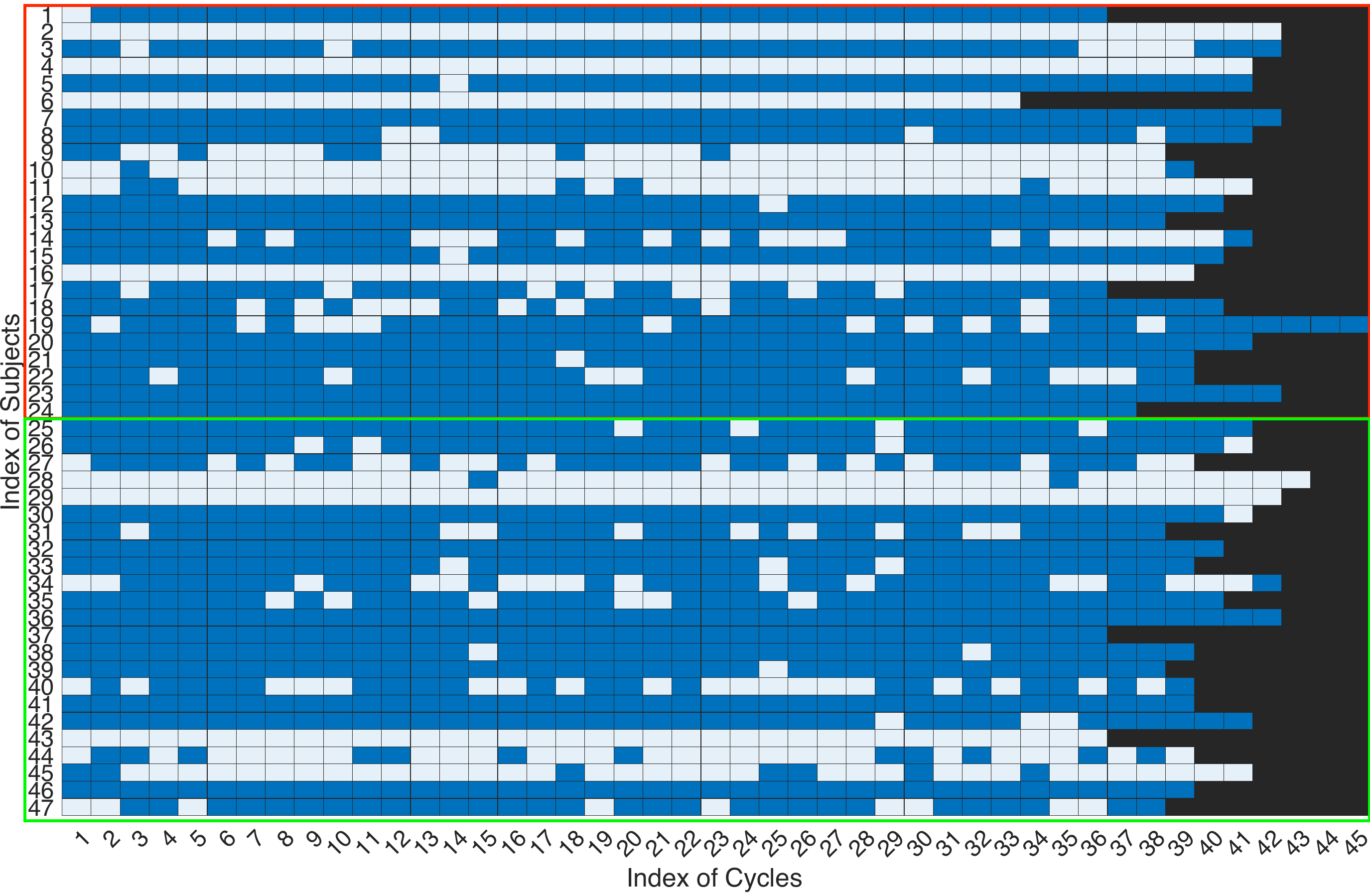}}
    \subfigure[KSSC]{
    \includegraphics[width=0.48\textwidth]{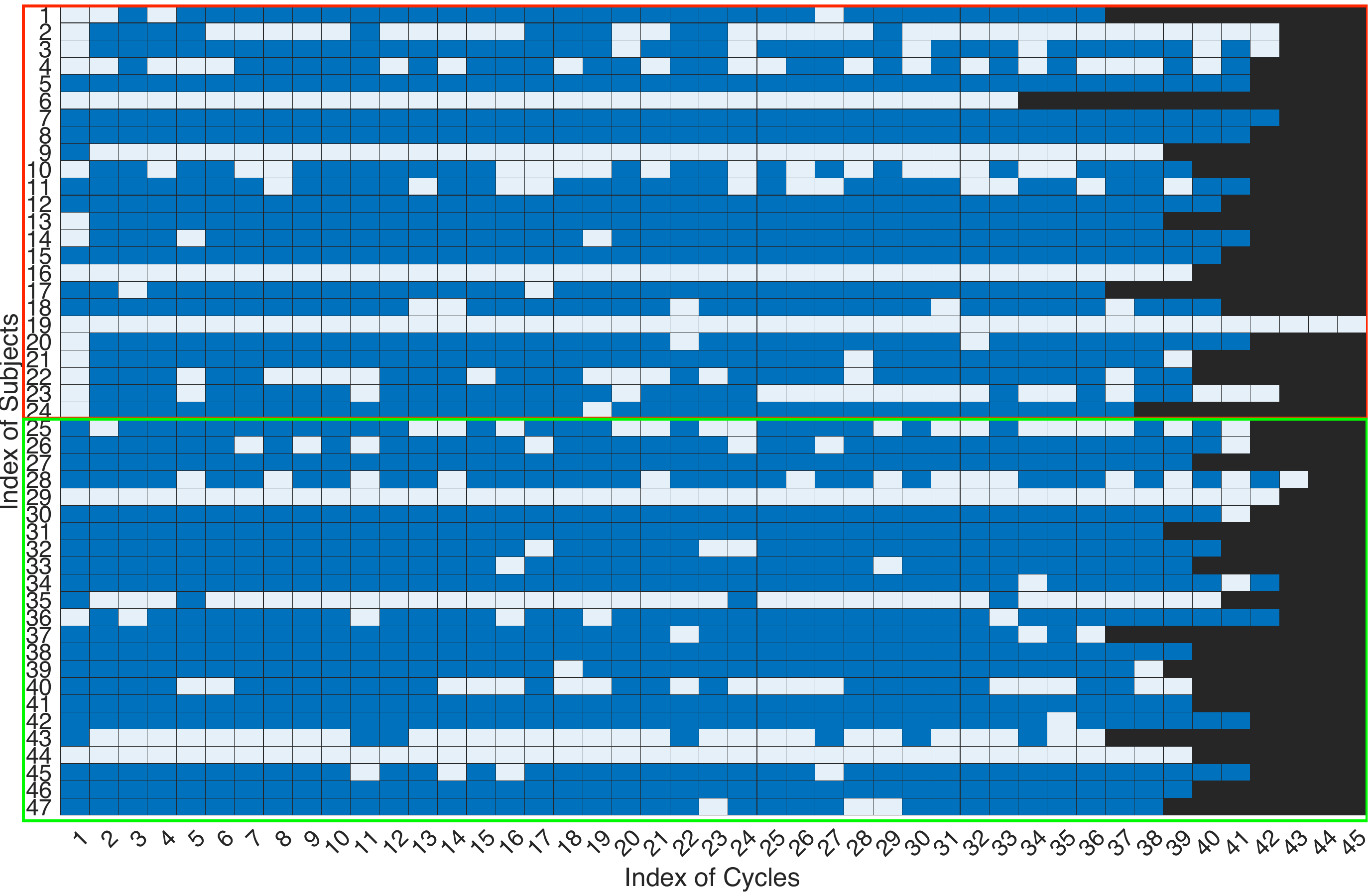}}
    \caption{A visualization of classification results for all subjects using the best subspace based on Channel 1 (sagittal plane of right hip). CAI subjects are framed in red and control subjects are framed in green. Correct predicted cycle is colored in blue whereas wrong predicted cycle is colored in white. Note that each subject has different number of running cycles, so no data are colored in black in the last cycles.}
    \label{heat_s}
\end{figure*}


In Figure \ref{heat_s}, we illustrate the classification results of both SSC and KSSC methods for all subjects are reported based on the best subspace of Channel 1 (sagittal plane of right hip). Correct prediction is colored in blue, wrong prediction is colored in white, and no data entry is colored in black. Subjects with CAI are framed in red and control subjects are framed in green. As observed, with SSC features, we classified running cycles correctly for most subjects except CAI Subjects 2, 4, 6, 9, 10, 11, and 16, and control Subjects 28, 29, 43, 44, and 45. With KSSC features, we classified running cycles correctly for most subjects except for CAI Subjects 2, 6, 9, 16, and 19, and control Subjects 9, 35, 43, and 44.

We further compare classification performance of our proposed SSC and KSSC methods to other statistical and PCA based extraction methods. We compute statistic features: mean and variance of motion time series. Pearson correlation is calculated to measure the linear relationships for all pairs of two motion time series and a correlation matrix is formed. PCA, a classical dimensionality reduction technique, is used to map a set of motion time series into a set of uncorrelated time series (called principal components) in a new space. We consider the top three principal components that account for 99.9 percent of variance in the original data space. Then we calculate the mean and variance of principal components to form a feature vector. Table \ref{Result Table} presents all experimental results and shows that our proposed SSC and KSSC methods outperforms other methods by $> 15\%$.

\begin{table}[ht]
   \centering
   \caption{Classification performance comparison of our proposed SSC and KSSC methods with statistical-based and PCA-based feature extraction methods. The average hit rates (upper panel) and majority voting accuracies (lower panel) are reported, respectively, from leave-one-subject-out cross validation. Each subject contains $\sim$40 sample cycles.}
   \vspace{0.1in}
    \begin{tabular}{|c|c|c|c|c|c|}\hline
    \centering
         Feature Learning & Statistical & Correlation & PCA & SSC & KSSC\\ \hline
        Testing Hit Rate & 57.49\% & 56.72\% & 52.25\% & \textbf{68.50}\% & \textbf{72.16}\% \\ \hline
        Training Hit Rate & 85.23\% & 77.73\% & 70.53\% & 76.13\% & 83.13\% \\ \hline \hline 
            Feature Learning & Statistic & Correlation & PCA & SSC & KSSC\\ \hline
        Testing Voting Accuracy & 61.70\% & 57.45\% & 51.06\% & \textbf{74.47}\% & \textbf{80.85}\% \\ \hline
        Training Voting Accuracy & 91.91\% & 79.56\% & 74.42\% & 75.53\% & 86.36\% \\ \hline
    
    \end{tabular}
\label{Result Table}
\end{table}

Our experiments also show that KSSC method outperforms SSC method. It means motion patterns are more discriminative on in a higher dimensional (non-linear) space, but not the original space. However, direct inter-dependency between joints /channels in motion patterns becomes not easily interpretable since all channels are tangle with each other. 

\subsection{Results for Network-based Subspace}
\label{network}

Consider the multi-joint coordinate system as a network of 18 nodes (representing channels). We use the network coefficients $\mathbf{C}_{s}$, computed in Section \ref{DBSCAN}, to describe the connectivity pattern among channels. We preform DBSCAN on every cycle for all subjects and obtained resulting clusters for every cycle. Then, to obtain the aggregating cluster information for all subjects, we consider two channels in the same subspace only if more than 50\% of subjects cluster these two channels into the same subspace. It turns out that two representative SSC-based clusters are formed in our experiment. Cluster I, resulting in higher classification accuracy, includes Channels 1, 2, 5, 7, 10, 11, 16, 17 and 18, for which Channels 1, 2 and 6 locate at hip joints, Channels 7, 10 and 11 locate at knee joints, and Channels 16, 17 and 18 locate at left ankle joint. The remaining channels are included in Cluster II. We then solve the SSC problem to obtain new network coefficients for Clusters I and II, which are used for pattern classification task. Note that we do not consider KSSC method because of burdensome interpretation of inter-dependency among channels.

Table \ref{subcluster} presents the testing results (hit rate and majority voting accuracy) for network-based subspace. We compare our method to other clustering methods based on human physical joints (hips, knees, and ankles) and Pearson correlation, as well as the baseline (including all channels in a cluster). Using the same DBSCAN setting, two Pearson correlation-based clusters are found. For all resultant clusters, we employ the SSC method for the channels in the clusters to obtain network coefficients $\mathbf{C}_{s}$. The subspace (Cluster I) learnt by our SSC method results in the highest accuracy (68\%). Between two correlation-based clusters, Cluster II results in a higher accuracy (65\%) and includes Channels 3, 4, 5, 6, 8, 11, 13, 15 and 16, for which most channels locate on hips and ankles. If we subjectively select the joints for testing, we find that hip joint is the best cluster compared to knees and ankles.

\begin{figure}[htbp]
    \centering
    \subfigure[Weights in SVM for SSC-based Cluster I]{\includegraphics[width=0.49\textwidth]{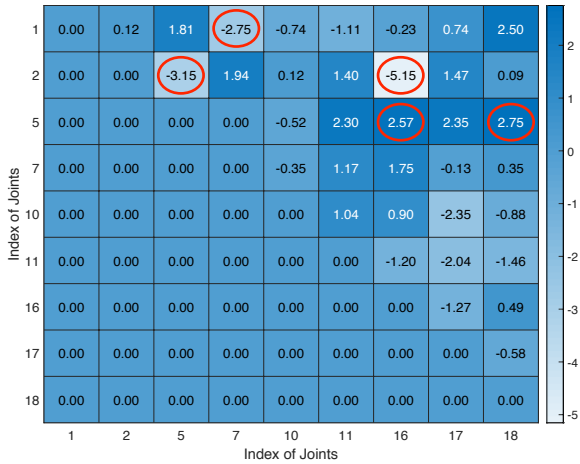}}
    \subfigure[Weights in SVM for correlation-based Cluster II]{\includegraphics[width=0.49\textwidth]{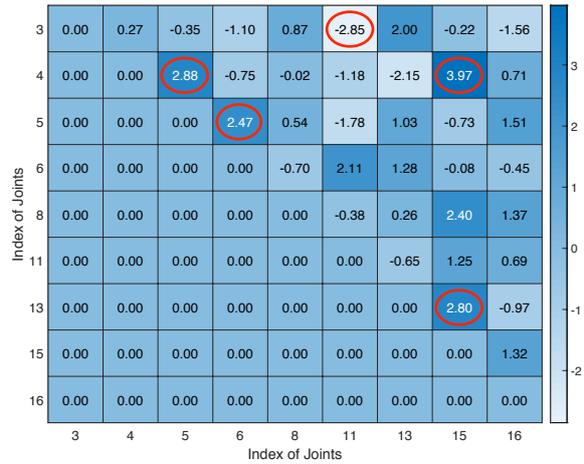}}
    \caption{A presentation of weights in SVM for both SSC-based Cluster I and correlation-based Cluster II. All subjects are included in the SVM model training. The five highest weights are marked in red circles regardless of sign.}
    \label{svmweight}
\end{figure}

In addition, it is noticed that the testing accuracy (68.09\%) from network-based algorithm is lower than one (74.47\%)  from channel-based algorithm. It might be because of some information loss from considering the only channels in the sub-cluster. However, for network-based subspace learning, we intended to observe the inter-dependency between channels in the representative subspace. In Figure \ref{svmweight}, we present the weights of SVM model using all subjects as training dataset for both correlation-based Cluster II and SSC-based Cluster I. As known, in SVM, features with higher weights (regardless of sign) are more important for classification. Inter-dependency corresponding to higher weights are found between channels in hip and ankle joints. In summary, recognizing motion (running) abnormality for CAI may be diagnosed with hip joints rather than ankle joints. It is suspected that subjects with CAI in our study may use muscles around hips to adjust their posture to protect their ankle from sprains.

\begin{table}[htbp]
    \centering \small  
    \caption{A testing performance comparison of network-based subspaces by the SSC methods and human physical joints (hips, knees, and ankles) and Pearson correlation, as well as the baseline.}
    \vspace{0.1in}
    \begin{tabular}{|c|c|c|c|c|c|c|c|c|}\hline
    \multirow{2}{*}{} & 
    \multirow{2}{*}{Baseline} &
    \multicolumn{3}{c|}{Physical joints} & \multicolumn{2}{c|}{Correlation based} & \multicolumn{2}{c|}{SSC based} \\
    \cline{3-9}
         ~ & ~ & Hips & Knees & Ankles & Cluster I & Cluster II & Cluster I & Cluster II \\ \hline
         Hit rate & 50.69\% & 60.53\% & 43.40\% & 46.09\% & 50.21\% & 60.15\% & \bf{67.37\%} & 50.04\% \\
         \hline
         $MV\_Accur$ & 55.32\% & 65.96\% & 44.68\% & 46.81\% & 48.94\% & 65.96\% & \bf{68.09\%} & 51.06\% \\ \hline
    \end{tabular}
    \label{subcluster}
\end{table}



\section{Conclusion} 
\label{sec5}
In this work, we studied a physical injury problem caused by ankle sprains, which develops a long-term and chronic instability in human daily movements. While most of recent studies were focused on the application of traditional statistic methods to biomechanical gait data, in this research we proposed a new analytic method to quantitatively learn the functional subspace that represents the original multivariate gait motion data acquired from the 3D motion capture system. We expected that there are joint effects as a network system, i.e., a multi-joint coordinate system of bilateral ankles, knees, and hips. Our proposed sparse subspace clustering algorithms attempted to learn significant subspaces in a lower dimensional space such that original motion dynamics can be easily characterized. A SVM classification model was trained with the extracted network-based features and validated for subjects with CAI compared to control subjects. We obtained higher classification performances by $>$10\% compared to the statistic-based features (that resulted in 55-60\% accuracy). Moreover, we found that there indeed exist joint effects among these channels from our analysis. The motion instability caused by ankle sprains are found significantly on the hip joints in the studied subjects. Subjects with CAI might use hip to stabilize and balance during movement. We have shown a potential that this proposed model can be applied to support the decisions in diagnosis, treatment, and rehabilitation. In this study, we extracted features from individual cycles, which may contain only local segment information. For future work, we will investigate global motion behaviors by extracting dynamic features from multiple running cycles in a sequence in the experiment, which may improve the pattern classification performance. We will further investigate abnormal detection problem - when motion behaviors begin to deteriorate during movement in individual subjects with CAI. Further, a dynamic modeling approach is suggested to obtain a functional clustering tensor. Given that our current study does not consider the geometry structure of multi-varate gait data, incorporating  geometry information may improve the understanding of the motion patterns in the human body system. 



\section*{Acknowledgement} 
This work is supported by Northeastern TIER 1 Seed Grant Program. 

\onehalfspacing

\end{document}